\pdfoutput=1

\PassOptionsToPackage{table,dvipsnames}{xcolor}

\documentclass[11pt]{article}

\usepackage{acl}

\usepackage{times}
\usepackage{latexsym}

\usepackage[T1]{fontenc}

\usepackage[utf8]{inputenc}

\usepackage{microtype}

\usepackage{inconsolata}

\usepackage{listings}
\usepackage{graphicx}
\usepackage{amsmath}
\usepackage{booktabs}
\usepackage{comment}
\usepackage{multirow}
\usepackage{fdsymbol} 
\usepackage[normalem]{ulem} 
\useunder{\uline}{\ul}{}
\usepackage{tabularray}

\lstdefinestyle{jsonstyle}{
    basicstyle=\ttfamily\footnotesize,
    stringstyle=\color{red},
    stepnumber=1,
    showstringspaces=false,
    breaklines=true,
    frame=single,
    backgroundcolor=\color{Gray!30},
    string=[s]{"}{"},
    comment=[l]{:\ "},
    morecomment=[l]{:"},
    literate=
     *{0}{{{\color{blue}0}}}{1}
      {1}{{{\color{blue}1}}}{1}
      {2}{{{\color{blue}2}}}{1}
      {3}{{{\color{blue}3}}}{1}
      {4}{{{\color{blue}4}}}{1}
      {5}{{{\color{blue}5}}}{1}
      {6}{{{\color{blue}6}}}{1}
      {7}{{{\color{blue}7}}}{1}
      {8}{{{\color{blue}8}}}{1}
      {9}{{{\color{blue}9}}}{1}
      {:}{{{\color{purple}:}}}{1}
      {,}{{{\color{purple},}}}{1}
      {\{}{{{\color{orange}\{}}}{1}
      {\}}{{{\color{orange}\}}}}{1}
      {[}{{{\color{orange}[}}}{1}
      {]}{{{\color{orange}]}}}{1},
}
\lstdefinestyle{textstyle}{
    basicstyle=\ttfamily\footnotesize,
    stringstyle=\color{Cerulean},
    stepnumber=1,
    showstringspaces=false,
    breaklines=true,
    backgroundcolor=\color{GreenYellow!15},
    string=[s]{\#}{:},
    comment=[l]{:\ "},
    tabsize=1,
    morecomment=[l]{:"},
    literate=
     *{0}{{{\color{blue}0}}}{1}
      {1}{{{\color{blue}1}}}{1}
      {2}{{{\color{blue}2}}}{1}
      {3}{{{\color{blue}3}}}{1}
      {4}{{{\color{blue}4}}}{1}
      {5}{{{\color{blue}5}}}{1}
      {6}{{{\color{blue}6}}}{1}
      {7}{{{\color{blue}7}}}{1}
      {8}{{{\color{blue}8}}}{1}
      {9}{{{\color{blue}9}}}{1}
      {:}{{{\color{purple}:}}}{1}
      {,}{{{\color{purple},}}}{1}
      {\{}{{{\color{orange}\{}}}{1}
      {\}}{{{\color{orange}\}}}}{1}
      {[}{{{\color{orange}[}}}{1}
      {]}{{{\color{orange}]}}}{1},
}
\lstdefinestyle{reviews}{
    basicstyle=\ttfamily\footnotesize,
    stepnumber=1,
    showstringspaces=false,
    breaklines=true,
    frame=single,
    backgroundcolor=\color{GreenYellow!15},
    string=[s]{"}{"},
    stringstyle=\color{red},
    comment=[s]{:\ "}{"},
    commentstyle=\color{green!50!black}
}

\title{eC-Tab2Text: Aspect-Based Text Generation from e-Commerce \\Product Tables}

\author{
Luis Antonio Gutiérrez Guanilo$^{\clubsuit}$ \enspace Mir Tafseer Nayeem$^{\vardiamondsuit}$\thanks{\ Corresponding authors.}  \\ 
\bf Cristian López$^{\clubsuit}$ \enspace Davood Rafiei$^{\vardiamondsuit}$\footnotemark[1] \\
$^{\clubsuit}$University of Engineering and Technology (UTEC) \enspace 
$^{\vardiamondsuit}$University of Alberta\\
\texttt{\{mnayeem, drafiei\}@ualberta.ca} \enspace  \enspace \texttt{\{luis.gutierrez.g, clopezd\}@utec.edu.pe} \\
}

\begin{document}
\maketitle

\begin{abstract}
Large Language Models (LLMs) have demonstrated exceptional versatility across diverse domains, yet their application in e-commerce remains underexplored due to a lack of domain-specific datasets. To address this gap, we introduce \textbf{eC-Tab2Text}, a novel dataset designed to capture the intricacies of e-commerce, including detailed product attributes and user-specific queries. Leveraging eC-Tab2Text, we focus on text generation from product tables, enabling LLMs to produce high-quality, attribute-specific product reviews from structured tabular data. Fine-tuned models were rigorously evaluated using standard Table2Text metrics, alongside correctness, faithfulness, and fluency assessments. Our results demonstrate substantial improvements in generating contextually accurate reviews, highlighting the transformative potential of tailored datasets and fine-tuning methodologies in optimizing e-commerce workflows. This work highlights the potential of LLMs in e-commerce workflows and the essential role of domain-specific datasets in tailoring them to industry-specific challenges\footnote{Our code, dataset, evaluation, model outputs, and other resources are publicly available at \href{https://github.com/Luis-ntonio/eC-Tab2Text}{eC-Tab2Text}.
}.
\end{abstract}

\begin{figure}[t]
    \centering
    \includegraphics[scale = 0.56]{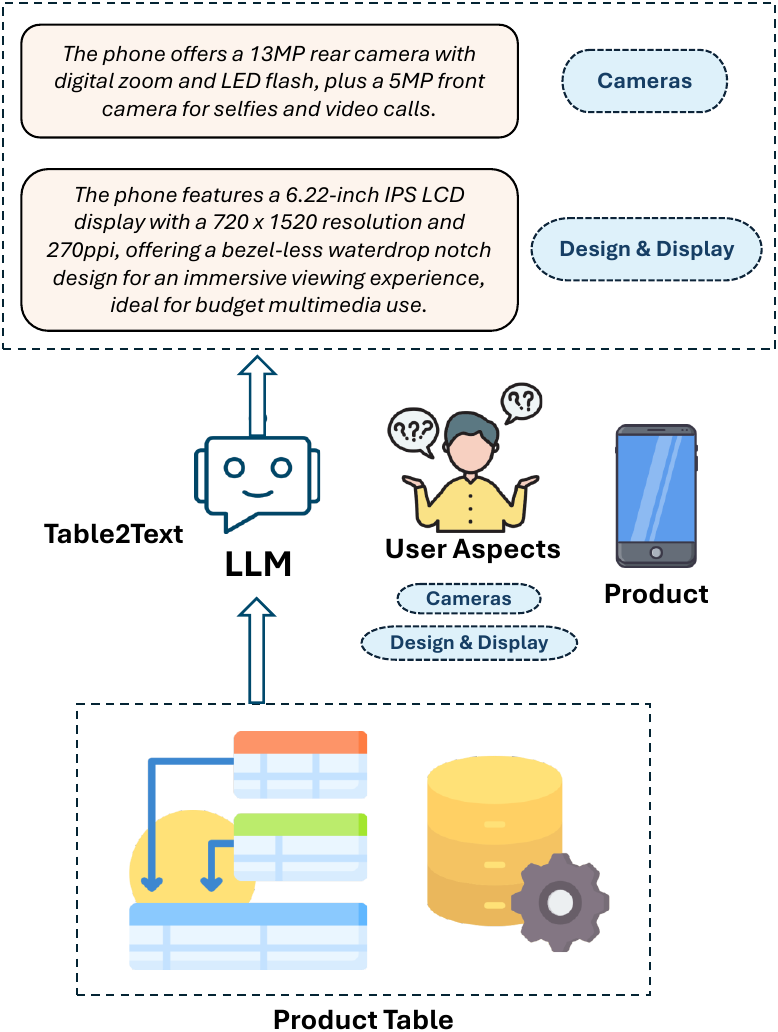}
    \caption{Overview of \textbf{eC-Tab2Text}. Illustration of aspect-based text generation from e-commerce product tables, where an LLM generates summaries for user-specific aspects like ``Camera'' and ``Design \& Display.''}
    \label{fig:task-def}
\end{figure}

\section{Introduction}
E-commerce relies heavily on tabular data, such as product details and features, while user interactions, including assistant agents and Q\&A, predominantly occur in natural language. This disparity underscores the need for models that can effectively parse tabular data and engage users through coherent, context-aware communication \cite{zhao-etal-2023-investigating}.
Table-to-text generation addresses this challenge by transforming structured data into natural language, enabling applications such as product reviews, personalized descriptions, and tailored summaries in e-commerce. Beyond e-commerce, this capacity extends to domains such as healthcare, where structured patient records are converted into concise summaries for doctors \cite{he2023survey}, and finance, where tabular financial data is transformed into analytical reports \cite{Varshney_2024}. However, generating text that is coherent, contextually relevant, and aligned with user-specific requirements remains a significant challenge, particularly for user- or query-centric tasks that demand domain-specific knowledge.
Existing table-to-text datasets often focus on general-purpose applications and lack the depth required for specialized domains. 
For instance, datasets like QTSumm \cite{zhao2023qtsummqueryfocusedsummarizationtabular} offer tabular summaries unrelated to the product domain, limiting their relevance for generating attribute-specific product reviews. E-commerce text generation requires handling diverse attributes (e.g., battery life, display quality), reasoning across different attributes (e.g., battery life and display size) and adapting to various user intents, such as crafting targeted product reviews \cite{macková2023promapdatasetsproductmapping}. 

\begin{table*}[ht]
    \footnotesize
    \centering
    \renewcommand{\arraystretch}{1.1} 
    \resizebox{16cm}{!} 
    { 
    \begin{tblr}{hline{1,2,Z} = 0.8pt, hline{3-Y} = 0.2pt,
                 colspec = {Q[l,m, 13em] Q[l,m, 6em] Q[c,m, 8em] Q[c,m, 5em] Q[l,m, 14em]},
                 colsep  = 4pt,
                 row{1}  = {0.4cm, font=\bfseries, bg=gray!30},
                 row{2-Z} = {0.2cm},
                 }
\textbf{Dataset}       & \textbf{Table Source} & \textbf{\# Tables / Statements} & \textbf{\# Words / Statement} & \textbf{Explicit Control}\\ 
\SetCell[c=5]{c} \textit{Single-sentence Table-to-Text}\\
ToTTo \cite{parikh2020tottocontrolledtabletotextgeneration}   & Wikipedia        & 83,141 / 83,141                  & 17.4                          & Table region      \\
LOGICNLG \cite{chen2020logicalnaturallanguagegeneration} & Wikipedia        & 7,392 / 36,960                  & 14.2                          & Table regions      \\ 
HiTab \cite{cheng-etal-2022-hitab}   & Statistics web   & 3,597 / 10,672                  & 16.4                          & Table regions \& reasoning operator \\ 
\SetCell[c=5]{c} \textit{Generic Table Summarization}\\
ROTOWIRE \cite{wiseman2017challengesdatatodocumentgeneration} & NBA games      & 4,953 / 4,953                   & 337.1                         & \textbf{\textit{X}}                   \\
SciGen \cite{moosavi2021scigen} & Sci-Paper      & 1,338 / 1,338                   & 116.0                         & \textbf{\textit{X}}                   \\
NumericNLG \cite{suadaa-etal-2021-towards} & Sci-Paper   & 1,355 / 1,355                   & 94.2                          & \textbf{\textit{X}}                    \\
\SetCell[c=5]{c} \textit{Table Question Answering}\\
FeTaQA \cite{nan2021fetaqafreeformtablequestion}     & Wikipedia      & 10,330 / 10,330                 & 18.9                          & Queries rewritten from ToTTo \\
\SetCell[c=5]{c} \textit{Query-Focused Table Summarization}\\
QTSumm \cite{zhao2023qtsummqueryfocusedsummarizationtabular}                        & Wikipedia      & 2,934 / 7,111                   & 68.0                          & Queries from real-world scenarios\\ 
\textbf{eC-Tab2Text} (\textit{ours})                           & e-Commerce products      & 1,452 / 3,354                   & 56.61                          & Queries from e-commerce products\\
    \end{tblr}
    }
\caption{Comparison between \textbf{eC-Tab2Text} (\textit{ours}) and existing table-to-text generation datasets. Statements and queries are used interchangeably. Our dataset specifically comprises tables from the e-commerce domain.}
\label{tab:datasets}
\end{table*}

While Large Language Models (LLMs) excel in general-purpose text generation \cite{touvron2023llama, kabir-etal-2024-benllm}, and fine-tuned models like \texttt{LLama2} \cite{touvron2023llama}, resulting in \texttt{StructLM} \cite{zhuang2024structlm} have shown improved performance on table-based datasets, these approaches often struggle with the complexities of product-specific domains. Addressing these intricacies requires tailored datasets to capture the nuanced requirements of attribute-specific text generation. Table-to-text generation has benefited from datasets like WikiTableT \cite{chen2021wikitabletlargescaledatatotextdataset}, TabFact \cite{2019TabFactA}, and ROTOWIRE \cite{wiseman2017challengesdatatodocumentgeneration}.
However, these datasets, designed for tasks like Wikipedia table descriptions, fact-checking, and sports summaries, lack the relevance for product-specific applications. Similarly,  LogicNLG \cite{chen2020logicalnaturallanguagegeneration} and ToTTo \cite{parikh2020tottocontrolledtabletotextgeneration} emphasize logical inferences and refined sentence extraction but fall short in addressing the demands of e-commerce text generation \cite{He2023ReviewOS}.

This paper introduces a tailored table-to-text dataset for the products domain and explores the potential of fine-tuned LLMs to bridge the gap between general-purpose capabilities and domain-specific needs. By leveraging domain-specific datasets and fine-tuning techniques, this work aims to empower e-commerce platforms to generate more precise and engaging product reviews given user aspects and tables (see Figure \ref{fig:task-def}), enhancing customer satisfaction and business outcomes.

Our main contributions are as follows: 

\begin{itemize}
    \item We present \textbf{eC-Tab2Text}, a novel domain-specific dataset for table-to-text generation in the e-commerce domain. The dataset features attribute-rich product tables paired with user-specific queries and outputs.
        
    \item We fine-tune open-source LLMs on the \textbf{eC-Tab2Text} dataset, resulting in significant improvements in text generation performance across various metrics. 
        
    \item We provide a detailed analysis of domain robustness by comparing models trained on \textbf{eC-Tab2Text} with those trained on QTSumm, highlighting the critical need for domain-specific datasets to achieve superior performance in e-commerce applications.
\end{itemize}

\section{Related Work}

\paragraph{Table-to-Text Generation} 
Table-to-text generation has advanced through datasets tailored to diverse domains and applications, as summarized in Table \ref{tab:datasets}. Early efforts, such as WikiTableT \cite{chen2021wikitabletlargescaledatatotextdataset}, focused on generating natural language descriptions from Wikipedia tables, while TabFact \cite{2019TabFactA} introduced fact-checking capabilities and ROTOWIRE \cite{wiseman2017challengesdatatodocumentgeneration} generated detailed sports summaries. However, these datasets are limited in their relevance to product-specific domains. Later datasets like LogicNLG \cite{chen2020logicalnaturallanguagegeneration} emphasized logical inference and reasoning, and ToTTo \cite{parikh2020tottocontrolledtabletotextgeneration} supported controlled text generation by focusing on specific table regions. HiTab \cite{cheng-etal-2022-hitab} extended these capabilities with hierarchical table structures and reasoning operators. Despite these advancements, none of these datasets provide the contextual and attribute-specific depth necessary for e-commerce applications, where generating meaningful descriptions requires reasoning across heterogeneous attributes, such as linking battery capacity to battery life or associating display size with user experience.

\paragraph{Query-Focused Summarization (QFS)} 
Advances in text summarization have improved multi-document summarization through abstractive methods like paraphrastic fusion \cite{10.1145/3132847.3133106, nayeem-etal-2018-abstractive}, compression \cite{10.1007/978-3-030-15719-7_14, chowdhury-etal-2021-unsupervised}, and diverse fusion models \cite{FUAD2019216, nayeem2017methods}, among others \cite{nayeem-chali-2017-extract, chali-etal-2017-towards}. These approaches lay the groundwork for query-focused summarization (QFS), which tailors summaries to user-specific queries. Initially formulated as a document summarization task, QFS aims to generate summaries tailored to specific user queries \cite{dang-2006-duc}. Despite its potential real-world applications, QFS remains a challenging task due to the lack of datasets. In the textual domain, QFS has been explored in multi-document settings \cite{giorgi-etal-2023-open} and meeting summarization \cite{zhong-etal-2021-qmsum}. Recent datasets like QTSumm \cite{zhao2023qtsummqueryfocusedsummarizationtabular} extend QFS to a new modality, using tables as input. However, QTSumm's general-purpose nature limits its applicability to product reviews, which require nuanced reasoning over attributes and user-specific contexts. Additionally, its queries are often disconnected from real-world e-commerce scenarios. In contrast, our proposed dataset, \textbf{eC-Tab2Text}, bridges this gap by providing attribute-specific and query-driven summaries tailored to e-commerce product tables.

\section{eC-Tab2Text: Dataset Construction}
\label{sec:dataset-contruction}

\begin{figure*}[t] 
    \centering
    \includegraphics[scale = 0.64]{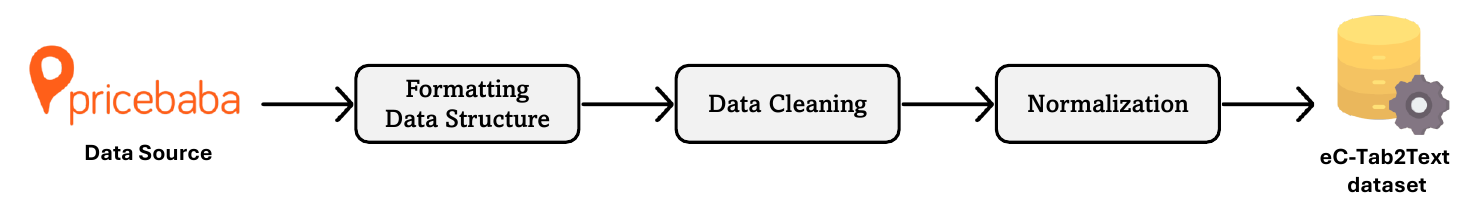} 
    \caption{Data collection pipeline for our \textbf{eC-Tab2Text} dataset.}
    \label{fig:data-pipeline} 
\end{figure*}

To address the gap in table-to-text generation for user-specific aspects or queries, such as ``Camera'' and ``Design \& Display'' (as illustrated in Figure \ref{fig:task-def}), we developed the \textbf{eC-Tab2Text} dataset. This dataset comprises e-commerce product tables and is designed to facilitate aspect-based text generation by fine-tuning LLMs on our dataset. The pipeline for creating \textbf{eC-Tab2Text} is outlined in Figure \ref{fig:data-pipeline} and described in detail below.

\paragraph{Data Sources} 
The dataset was constructed using product reviews and specifications (i.e., tables) extracted from the Pricebaba website\footnote{\url{https://pricebaba.com}, last accessed August 2024.}. Pricebaba provides comprehensive information on electronic products, including mobile phones and laptops. For this study, the focus was exclusively on mobile phone data due to the richness of product specifications (attribute-value pairs) and the availability of detailed expert reviews as summaries. Additionally, the number of samples available for mobile phones is significantly larger than for laptops. Each sample includes feature-specific details such as camera performance, battery life, and display quality.

\paragraph{Data Extraction and Format} 
Data extraction was performed using web scraping techniques, with the extracted data stored in JSON format to serialize the table structure and to ensure compatibility with modern data processing workflows. Two JSON files were generated (Appendix \ref{Appendix: eC-Tab2Text Data Collection}): one containing aspect-based product reviews and the other containing product specifications. The review JSON file captures user aspects alongside their associated textual descriptions collected from the ``Quick Review'' section of the website, while the specifications JSON file stores key-value pairs for both key specifications and full technical details. The structures of the sample inputs and outputs are depicted in Figures \ref{fig:pricebaba-review-structure} and \ref{fig:pricebaba-spec-structure} in the Appendix.

\paragraph{Data Cleaning, Normalization, and Integration} 
To ensure consistency, usability, and completeness, the extracted data underwent rigorous cleaning, normalization, and integration, similar to previous approaches \cite{nayeem-rafiei-2023-role, nayeem-rafiei-2024-kidlm, nayeem2024lfosum}. The process includes \textbf{(1)} standardizing all text values to lowercase for uniformity, \textbf{(2)} replacing special characters (e.g., \texttt{\&} with ``and'') to improve readability, and \textbf{(3)} normalizing keys to maintain logical and contextual coherence.
For example, the key \texttt{Display \& Design} was transformed into \texttt{Design and Display} to improve readability and alignment with naming conventions.

To further enhance the dataset quality, irrelevant and redundant entries were removed through a systematic filtering process: \textbf{(1)} reviews lacking textual content in the text field were discarded, \textbf{(2)} specifications containing only generic or minimal information (e.g., entries labeled as  \texttt{General}) were excluded, \textbf{(3)} overly simplistic reviews categorized as \texttt{Overview} were omited to maintain a focus on detailed and meaningful content.

Finally, the reviews and specifications JSON files were merged into a unified dataset by aligning entries based on their unique product URLs. This integration consolidated each product's reviews and specifications into a single, cohesive record, creating a streamlined and comprehensive dataset for downstream applications.

\begin{table}[!ht]
    \footnotesize
    \centering
    \begin{tblr}{hline{1,2,Z} = 0.8pt, hline{3-Y} = 0.2pt,
                 colspec = {Q[l,m, 13em] Q[l,m, 6em]},
                 colsep  = 4pt,
                 row{1}  = {0.4cm, font=\bfseries, bg=gray!30},
                 row{2-Z} = {0.1cm},
                 }
    \textbf{Metric}       & \textbf{Value} \\ 
    \SetCell[c=2]{c} \textit{Input}\\
    \# Tables & 1,452\\
    Avg \# Attribute-Value Pairs & 59.8\\
    Max \# Attribute-Value Pairs & 68\\
    \SetCell[c=2]{c} \textit{Output}\\
    \# Queries & 3,354\\
    Avg \# queries/table & 2.31\\
    Avg \# words/query & 56.61\\
    \end{tblr}
\caption{Statistics of our \textbf{eC-Tab2Text} dataset.}
\label{table:eC-Tab2Text-statistics}
\end{table}


Our \textbf{eC-Tab2Text} dataset provides a comprehensive resource for table-to-text generation tasks based on user queries, as summarized in Table~\ref{table:eC-Tab2Text-statistics}. The input JSON files contain attribute-rich product specifications, averaging 59.8 attribute-value pairs per table, with the largest entries containing up to 68 pairs. The dataset includes 3,354 queries, averaging 2.31 queries per table, with concise outputs averaging 56.61 words per query. This design supports query-specific training and evaluation of LLMs, enabling precise and contextually relevant text generation tailored to user queries.

\section{eC-Tab2Text: Models}
\label{sec:models}

This section outlines the methodology for table serialization and provides details on the selection and fine-tuning of LLMs using our dataset.

\paragraph{Table Serialization} 
The representation of tabular data in machine learning has been addressed through various serialization techniques, including markdown format, comma-separated values (CSV), HTML \cite{fang2024large, singha2023tabular}, and LaTeX \cite{jaitly2023betterserializationtabulardata}. However, for our specific problem involving semi-structured tables with nested structures, we adopt JSON serialization. This approach effectively addresses two critical needs: \textbf{(1)} representing the nested structures inherent in product tables and \textbf{(2)} enabling query-specific generation and evaluation \cite{gao2024jsontuninggeneralizablerobustcontrollable}.

In our eC-Tab2Text dataset, both input tables and query-specific outputs are serialized using JSON. The input JSON captures structured product specifications, while the output JSON aligns queries (e.g., ``Design and Display'' or ``Battery'') as keys and their corresponding generated texts as values. This unified representation facilitates efficient querying and maintains alignment between inputs and outputs, ensuring consistency across the dataset. Additional implementation details can be found in Appendix \ref{appendix:fine-tuning-models} (Listing \ref{code:Prompt Dataset 1} prompt).

\paragraph{Model Selection and Characteristics} 
To evaluate the effectiveness of the eC-Tab2Text dataset, we fine-tuned three open-source LLMs: \textbf{LLaMA 2-Chat 7B} \cite{touvron2023llama}, \textbf{Mistral 7B-Instruct} \cite{jiang2023mistral}, and \textbf{StructLM 7B} \cite{zhuang2024structlm}. These models were selected due to their distinct pretraining paradigms, which address diverse data modalities and tasks. Detailed descriptions of these models are provided in Appendix \ref{sec:fine-tuning-models} and summarized below.

\begin{itemize}
    \item \textbf{LLaMA 2-Chat 7B}\footnote{\href{https://huggingface.co/meta-llama/Llama-2-7b-chat-hf}{Llama-2-7b-chat-hf}}: This model, pretrained on 2 trillion tokens of publicly available text data, is fine-tuned on over one million human-annotated examples. It excels in general-purpose conversational and language understanding tasks \cite{touvron2023llama}.
    \item \textbf{Mistral 7B-Instruct}\footnote{\href{https://huggingface.co/mistralai/Mistral-7B-Instruct-v0.3}{Mistral-7B-Instruct-v0.3}}: Leveraging a mix of text and code during training, this model demonstrates strong performance in tasks that require natural language understanding and programming-related reasoning \cite{jiang2023mistral}.
    \item \textbf{StructLM 7B}\footnote{\href{https://huggingface.co/TIGER-Lab/StructLM-7B}{StructLM-7B }}: Pretrained on structured data, including databases, tables, and knowledge graphs, StructLM is optimized for structured knowledge grounding, making it particularly effective for domain-specific tasks \cite{zhuang2024structlm}.
\end{itemize}

\paragraph{Fine-Tuning Process} 
The fine-tuning process adapts these models to the e-commerce domain using the eC-Tab2Text dataset. This dataset focuses on attribute-specific and context-aware text generation tailored to user queries, such as detailed reviews of ``Camera'' or ``Design \& Display.'' The fine-tuning process follows best practices in instruction tuning and domain-specific dataset alignment \cite{Zhang2023InstructionTF, Chang2023ASO}. Optimization of hyperparameters ensured computational efficiency while maintaining high-quality performance, as detailed Table~\ref{table:hyperparameters}. 

By leveraging these diverse models and aligning them with the eC-Tab2Text dataset, this work aims to advance the state-of-the-art in domain-specific language generation for e-commerce applications.

\begin{table}[ht] 
\footnotesize 
\centering 
\begin{tblr}{hline{1,2,Z} = 0.8pt, hline{3-Y} = 0.2pt, colspec = {Q[c,m, 4.5em] Q[c,m, 4.5em] Q[c,m, 7em]}, colsep = 4pt, row{1} = {0.6cm, font=\bfseries, bg=gray!30}, row{2-Z} = {0.3cm}, }

\textbf{Column Name} & \textbf{Data Type} & \textbf{Description}\\ 
\texttt{table} & Dictionary & Contains structured data with headers and rows.\\ 
\texttt{example\_id} & String & Unique identifier for each dataset example.\\ 
\texttt{query} & String & Textual description or query related to the dataset.\\ 
\texttt{summary} & String & Summary or explanation generated in response to the query.\\ 
\texttt{row\_ids} & Sequence of Integers & Row indices corresponding to the data referenced in the \texttt{table} column.\\ 
\end{tblr} 
\caption{Structure of the QTSUMM Dataset.}
\label{tab:qtsumm_structure} 
\end{table}
\begin{table}[ht]
\centering
\footnotesize
\begin{tabular}{|c|c|}
\hline
\textbf{Hyperparameter} & \textbf{Value} \\
\hline
Learning Rate & $2 \times 10^{-4}$ \\
Batch Size & 2 \\
Epochs & 1 \\
Gradient Accumulation Steps & 1 \\
Weight Decay & 0.001 \\
Max Sequence Length & 900 \\
\hline
\end{tabular}
\caption{Hyperparameter settings for fine-tuning.}
\label{table:hyperparameters}
\end{table}
\begin{table}[!ht]
    \footnotesize
    \centering
    \begin{tblr}{hline{1,2,Z} = 0.8pt, hline{3-Y} = 0.2pt,
                 colspec = {Q[c,m, 14em] Q[c,m, 6em]},
                 colsep  = 4pt,
                 row{1}  = {0.6cm, font=\bfseries, bg=gray!30},
                 row{2-Z} = {0.3cm},
                 }

\textbf{Hyperparameter} & \textbf{Value}      \\
\texttt{bnb\_4bit\_compute\_dtype}       & float16  \\
\texttt{bnb\_4bit\_quant\_type} & nf4               \\
\texttt{use\_nested\_quant}       & False           \\
    \end{tblr}
    \caption{Quantization settings used for fine-tuning.}
    \label{tab:eC-Tab2Text Aditional parameters}
\end{table}

\section{Evaluation} 
\label{sec:evaluation}

\begin{table*}[t]
\centering
\renewcommand{\arraystretch}{1.1} 
\resizebox{16cm}{!} 
{ 
\begin{tabular}{cccccccccc}
\hline
\rowcolor[HTML]{EFEFEF} 
\textbf{Mode} &
  \textbf{Models} &
  \multicolumn{1}{l}{\cellcolor[HTML]{EFEFEF}\textbf{BLEU}} &
  \textbf{METEOR} &
  \textbf{ROUGE-1} &
  \textbf{ROUGE-L} &
  \textbf{BERTScore} &
  \textbf{Correctness} &
  \textbf{Faithfulness} &
  \textbf{Fluency} \\ \hline \hline
                                      & \textbf{\texttt{Llama2}}           & 1.39  & 3.59  & 5.57  & 4.09  & 66.49 & 32.18 & 37.68 & 32.47 \\
                                      & \textbf{\texttt{StructLM}}         & 6.21  & 11.96 & 20.09 & 15.34 & 82.56 & 64.30 & 70.08 & 63.10 \\
                                      & \textbf{\texttt{Mistral}}          & 4.19  & 9.55  & 25.64 & 18.99 & 82.12 & 77.02 & 81.16 & 76.5  \\
                                      & \textbf{\texttt{GPT-4o-mini}}      & 7.14  & 16.12 & 29.44 & 19.47 & 83.75 & \textbf{80.89} & \textbf{83.92} & \textbf{80.81} \\
\multirow{-5}{*}{\textbf{Zero-Shot}}       & \textbf{\texttt{Gemini-1.5-flash}} & 8.8   & 15.18 & 30.38 & 21.51 & 84.05 & {\ul 78.79} & {\ul 83.04} & {\ul 78.54} \\ \hline
                                      & \textbf{\texttt{Llama2}}           & 29.36 & 40.2  & 48.36 & 39.25 & 90.05 & 61.38 & 63.78 & 61.47 \\
                                      & \textbf{\texttt{StructLM}}         & {\ul 31.06} & {\ul 42.3}  & {\ul 49.42} & {\ul 40.58} & {\ul 90.9}  & 69.70 & 72.46 & 69.93 \\
\multirow{-3}{*}{\textbf{Fine-tuned}} & \textbf{\texttt{Mistral}}          & \textbf{38.89} & \textbf{49.43} & \textbf{56.64} & \textbf{48.32} & \textbf{92.18} & 73.07 & 76.63 & 73.03 \\ \hline
\end{tabular}
}
\caption{Evaluation results of zero-shot and fine-tuned models on the \textbf{eC-Tab2Text} dataset. The best results are highlighted in \textbf{bold}, and the second-best results are {\ul underlined}.}
\label{table:results1}
\end{table*}

\begin{table*}[t]
\centering
\renewcommand{\arraystretch}{1.1} 
\resizebox{16cm}{!} 
{ 
\begin{tabular}{ccccccccccc}
\hline
\rowcolor[HTML]{EFEFEF} 
\textbf{Dataset Trained} &
  \textbf{Dataset Tested} &
  \textbf{Models} &
  \textbf{BLEU} &
  \textbf{METEOR} &
  \textbf{ROUGE-1} &
  \textbf{ROUGE-L} &
  \textbf{BERTScore} &
  \textbf{Correctness} &
  \textbf{Faithfulness} &
  \textbf{Fluency} \\ \hline \hline
                                       &                                        & \textbf{\texttt{Llama2}}   & {\ul 13.32} & {\ul 32.38} & 26.3  & 19.22 & {\ul 86.47} & 51.09 & 57.30 & 48.98 \\
                                       &                                        & \textbf{\texttt{StructLM}} & 6.6   & 22.04 & 13.52 & 10.04 & 84.5  & 41.14 & 48.92 & 39.68 \\
                                       & \multirow{-3}{*}{\begin{tabular}[c]{@{}c@{}}\textbf{QTSumm}\\ (\texttt{In-domain})\end{tabular}}      & \textbf{\texttt{Mistral}}  & 10.1  & 28.57 & 20.7  & 15.51 & 85.65 & 49.99 & 57.73 & 50.71 \\ \cline{2-11} 
                                       &                                        & \textbf{\texttt{Llama2}}   & 17.47 & 40.2  & 35.69 & 21.14 & 85.41 & 63.98 & 71.40 & 64.07 \\
                                       &                                        & \textbf{\texttt{StructLM}} & 3.73  & 17.42 & 10.41 & 6.77  & 82.91 & 36.69 & 60.81 & 37.03 \\
\multirow{-6}{*}{\textbf{QTSumm}}      & \multirow{-3}{*}{\begin{tabular}[c]{@{}c@{}}\textbf{eC-Tab2Text}\\ (\texttt{Out-of-domain})\end{tabular}} & \textbf{\texttt{Mistral}}  & 13.97 & 26.88 & 28.58 & 17.08 & 84.83 & 58.35 & 69.81 & 58.95 \\ \hline \hline
                                       &                                        & \textbf{\texttt{Llama2}}   & 6.5   & 22.77 & 7.79  & 16.59 & 81.93 & 48.42 & 48.66 & 48.55 \\
                                       &                                        & \textbf{\texttt{StructLM}} & 10.15 & 30.59 & {\ul 30.59} & 23.04 & 85.13 & 58.71 & 56.60 & 58.26 \\
                                       & \multirow{-3}{*}{\begin{tabular}[c]{@{}c@{}}\textbf{QTSumm}\\ (\texttt{Out-of-domain})\end{tabular}}      & \textbf{\texttt{Mistral}}  & 10.39 & 18.11 & 30.27 & {\ul 24.24} & 84.23 & {\ul 64.83} & {\ul 61.14} & {\ul 64.51} \\ \cline{2-11} 
                                       &                                        & \textbf{\texttt{Llama2}}   & 29.4  & 40.21 & 48.43 & 39.25 & 90.05 & 61.38 & 63.78 & 61.47 \\
                                       &                                        & \textbf{\texttt{StructLM}} & 31.06 & 42.3  & 49.42 & 40.58 & 90.9  & 69.70 & 72.46 & 69.93 \\
\multirow{-6}{*}{\textbf{eC-Tab2Text}} & \multirow{-3}{*}{\begin{tabular}[c]{@{}c@{}}\textbf{eC-Tab2Text}\\ (\texttt{In-domain})\end{tabular}} & \textbf{\texttt{Mistral}}  & \textbf{38.89} & \textbf{49.43} & \textbf{56.64} & \textbf{48.32} & \textbf{92.18} & \textbf{73.07} & \textbf{76.63} & \textbf{73.03} \\ \hline
\end{tabular}
}
\caption{Robustness evaluation results on our \textbf{eC-Tab2Text} dataset and the QTSumm dataset \cite{zhao2023qtsummqueryfocusedsummarizationtabular}. The best results on our dataset, including both in-domain and out-of-domain scenarios, are highlighted in \textbf{bold}, while the best results on the QTSumm dataset, both in-domain and out-of-domain, are {\ul underlined}.}

\label{table:results}
\end{table*}

In this section, we evaluate the performance of the eC-Tab2Text models described in Section \ref{sec:models} along with several closed-source models, including GPT-4o-mini and Gemini-1.5-flash. The evaluation follows standard metrics commonly used in table-to-text generation, as outlined in \cite{zhao2023qtsummqueryfocusedsummarizationtabular}. These metrics include BLEU \cite{Reiter2018A}, the F-1 scores of ROUGE-1 and ROUGE-L \cite{Ganesan2015ROUGE}, METEOR \cite{Dobre2015ACB}, and BERTScore \cite{zhang2020bertscoreevaluatingtextgeneration}, following \cite{akash-etal-2023-shironaam, shohan-etal-2024-xl}. To assess the correctness, faithfulness, and fluency of the generated text, we employ PROMETHEUS 2 \cite{kim2024prometheus2opensource} and an open-source LLM-based evaluator as an alternative to the closed-source G-Eval \cite{liu2023gevalnlgevaluationusing}. Our objective is to benchmark the performance of various LLMs under both zero-shot and fine-tuned settings using the proposed eC-Tab2Text dataset.

\paragraph{Experimental Setup} The fine-tuning process was conducted on a NVIDIA RTX 4070 Ti Super GPU with 16GB of VRAM, ensuring efficient training while managing memory-intensive operations. The AdamW optimizer \cite{loshchilov2018decoupled} was configured with a learning rate of $2\times 10^{-4}$, chosen for its effectiveness in maintaining stability and convergence during training. To optimize resource usage, the \textit{bitsandbytes} library\footnote{\url{https://github.com/bitsandbytes-foundation/bitsandbytes}} was employed for 4-bit quantization, reducing VRAM requirements without significant performance loss. Table~\ref{tab:eC-Tab2Text Aditional parameters} outlines the key parameters used, including `\texttt{float16}' for computation data type and `\texttt{nf4}' for quantization type. The `\texttt{use\_nested\_quant}' option was set to `False' to ensure compatibility across models.

Detailed information on the evaluation metrics is included in Appendix \ref{sec:evaluation-metrics}. Our eC-Tab2Text dataset was divided into training and testing subsets, using an 80\%-20\% split. This ensures a sufficient volume of data for training while preserving a reliable subset for evaluation.

\subsection{Robustness Evaluation} 
We evaluate the robustness of the models under domain differences, focusing on their performance with in-domain and out-of-domain training data. The primary objective is to analyze how models perform when fine-tuned on data from different domains and to emphasize the importance of our proposed eC-Tab2Text dataset for the e-commerce product domain. For this evaluation, we compare the performance of models fine-tuned on the QTSumm dataset \cite{zhao2023qtsummqueryfocusedsummarizationtabular}, which contains Wikipedia tables with queries, against those fine-tuned on our eC-Tab2Text dataset, which consists of product tables with user-specific queries.

\paragraph{QTSumm Dataset Details} The QTSumm dataset, obtained from Hugging Face\footnote{\url{https://huggingface.co/datasets/yale-nlp/QTSumm}} provides structured data that facilitates query-specific text summarization tasks. The detailed structure of QTSumm is outlined in Table \ref{tab:qtsumm_structure}. This dataset's structure ensures a systematic alignment between the input queries, the corresponding structured data, and the generated summaries, making it a valuable benchmark for fine-tuning and evaluating the performance of LLMs in handling structured data. Its focus on query-specific summarization provided an excellent foundation for testing the robustness and adaptability of the proposed methodologies.

For fine-tuning, we utilized the same models described in Section \ref{sec:models}, employing identical hyperparameters. The models were trained using prompts structured consistently with those designed for the eC-Tab2Text dataset. However, in the QTSumm setup, the prompts included row-level content tailored to the dataset's structure, as outlined in Appendix \ref{appendix:fine-tuning-models} (Listing \ref{code:Prompt Dataset 2}). This alignment ensured methodological consistency while accounting for the unique characteristics of the QTSumm dataset. By highlighting these differences, our evaluation underscores the critical need for domain-specific datasets, such as eC-Tab2Text, to achieve robust and accurate performance in the 
product domain.

\subsection{Results \& Analysis}
Our experimental results, illustrated in Table~\ref{table:results1}, demonstrate that fine-tuning open-source 7B models on our dataset leads to substantial performance improvements. These fine-tuned models significantly outperform major proprietary models, such as GPT-4o-mini and Gemini-1.5-flash, across text-based metrics, including BLEU, ROUGE-1, ROUGE-L, METEOR, and BERTScore, while achieving competitive results in model-based metrics like faithfulness, correctness, and fluency, narrowing the gap with proprietary counterparts. This is significant given the relatively small size of our dataset compared to the much larger datasets used for training many proprietary models. Notably, Mistral\_Instruct, fine-tuned on our dataset, excels by achieving the highest scores across all metrics, surpassing both zero-shot and fine-tuned models.

As highlighted in Table~\ref{table:results}, the robustness of our dataset is further evidenced by comparing it against the QTSUMM dataset; models trained with our dataset consistently outperform those trained on QTSUMM across both in-domain and out-of-domain tasks, with Mistral\_Instruct leading, followed closely by StructLM. Although both datasets share similar task objectives, the domain differences significantly affect the models' performance. 

Outputs generated by different open-source models are presented in Mistral (Listing \ref{code:JSON-Mistral-eC-Tab2Text}), StructLM (Listing \ref{code:JSON-StructLM}), and Llama2 (Listing \ref{code:JSON-Llama2}), as well as by closed-source models GPT-4o-mini (Listing \ref{code:JSON-GPT4}) and Gemini1.5-flash (Listing \ref{code:JSON-Gemini}). Notably, the closed-source models tend to produce longer outputs compared to the open-source models, with their outputs often containing nested keys and detailed information.

\section{Discussion and Future Directions}
\label{sec:discussion-future}

This section highlights the need for better numerical reasoning in table-to-text generation and improved evaluation methods. 

\paragraph{Numerical Reasoning} Product tables, with their semi-structured and nested attributes (e.g., battery capacity in mAh, display size in inches), demand advanced numerical reasoning to generate meaningful text. Models must analyze relationships, such as how battery life depends on capacity and display size, or how display dimensions impact user experience. Unlike Wikipedia tables \cite{zhao2023qtsummqueryfocusedsummarizationtabular, nahid-rafiei-2024-tabsqlify}, which focus on factual text generation, our eC-Tab2Text dataset challenges models to integrate numerical reasoning with qualitative text generation \cite{islam-etal-2024-large}. This unique focus enables LLMs to synthesize structured data into nuanced, human-readable summaries, providing a benchmark for evaluating and improving reasoning capabilities in real-world applications \cite{10.1145/3583780.3615172, akhtar-etal-2023-exploring, zhao-etal-2024-docmath}. Future work could explore pushing the boundaries of LLMs capabilities in numerical and qualitative reasoning using our dataset.

\paragraph{Evaluation} Although we evaluated the correctness, faithfulness, and fluency of the generated text using PROMETHEUS 2 \cite{kim2024prometheus2opensource}, attribute-specific text evaluation against product tables requires a more nuanced approach. Future evaluations could involve extracting attribute-value pairs from the generated text \cite{shinzato-etal-2023-unified, brinkmann2024extractgptexploringpotentiallarge}, verifying their correctness and contextual relevance, and comparing them with the corresponding values in the source tables to enable more fine-grained and precise assessments.

\section{Conclusion}
This work introduces \textbf{eC-Tab2Text}, a novel dataset for table-to-text generation in the e-commerce domain, addressing the limitations of existing general-purpose datasets. By fine-tuning open-source LLMs, we demonstrate substantial improvements in generating attribute-specific, contextually accurate product reviews. Our evaluation highlights the robustness of \textbf{eC-Tab2Text}, outperforming comparable datasets like QTSumm, and underscores the importance of domain-specific datasets for advancing LLM performance in industry-specific applications. This study lays the groundwork for future research in expanding dataset scope, evaluation methodologies, and enhancing numerical reasoning in e-commerce workflows.

\section*{Limitations}
\label{sec:limitations}

In this work, we evaluated our proposed methods using a selection of both open-source and closed-source LLMs. We intentionally focused on cost-effective yet efficient closed-source models and open-source models deployable on consumer-grade hardware, considering the constraints of \emph{academic settings}. The performance of more powerful, large-scale models remains unexplored; however, we encourage the broader research community to benchmark these models using our dataset. To support future research, we make our code, dataset, evaluation, model outputs, and other resources publicly available\footnote{\url{https://github.com/Luis-ntonio/eC-Tab2Text}}.

This study faced several system and resource constraints that shaped the methodology and evaluation process. For example, VRAM limitations required capping the maximum token length at 900 for the Mistral\_Instruct model to ensure uniform hyperparameter settings across all models. While this standardization enabled consistent comparisons, it may have limited some models' ability to generate longer and potentially more nuanced outputs.

Our dataset focused exclusively on mobile phone data due to the richness of product specifications (attribute-value pairs) and the availability of detailed expert reviews as summaries. Future work could expand the dataset to include other domains, such as laptops, home appliances, and wearable devices, to assess the generalizability of the LLMs in e-Commerce domains. 

Finally, the development of eC-Tab2Text has been exclusively centered on the \textbf{English language}. As a result, its effectiveness and applicability may differ for other languages. Future research could explore multilingual extensions to broaden its usability across diverse linguistic and cultural contexts.

\section*{Ethics Statement}
 
The data scraping process for this research was conducted with strict adherence to ethical guidelines and solely for non-commercial research purposes, under the Creative Commons Attribution-NonCommercial-ShareAlike 4.0 International License (CC BY-NC-SA 4.0)\footnote{\url{https://creativecommons.org/licenses/by-nc-sa/4.0/}}. To minimize potential harm to the source website, measures were implemented to ensure controlled and responsible scraping practices. These safeguards were designed to avoid undue strain on the website’s infrastructure, such as preventing Distributed Denial of Service (DDoS) attacks, thereby maintaining the integrity and functionality of the site.

\section*{Acknowledgements}

We thank all the anonymous reviewers and the meta-reviewer for their valuable feedback and constructive suggestions. 
This research is supported by the Natural Sciences and Engineering Research Council of Canada (NSERC). Additionally, Luis Antonio Gutiérrez Guanilo is supported by the Emerging Leaders in the Americas Program (ELAP), and Mir Tafseer Nayeem is supported by a Huawei PhD Fellowship.

\bibliography{main}

\clearpage
\appendix
\twocolumn[{%
 \centering
 \Large\bf Supplementary Material: Appendices \\ [20pt]
}]

\section{Evaluation Metrics}
\label{sec:evaluation-metrics}


\begin{itemize} 
    \item \textbf{BLEU (Bilingual Evaluation Understudy)}: Commonly used in machine translation and natural language generation, BLEU measures the overlap of n-grams between generated and reference texts. Despite its popularity, BLEU has limitations, particularly in capturing semantic similarity and evaluating beyond exact matches \cite{Reiter2018A}. 
    \item \textbf{ROUGE (Recall-Oriented Understudy for Gisting Evaluation)}: Focuses on recall-oriented evaluation by comparing the overlap of n-grams, word sequences, and word pairs between generated summaries and reference texts. It is highly effective for summarization tasks \cite{Ganesan2015ROUGE}.
    \item \textbf{METEOR (Metric for Evaluation of Translation with Explicit ORdering)}: Incorporates stemming, synonymy, and flexible matching, providing a more nuanced evaluation than BLEU. It strongly correlates with human judgments, especially in translation tasks \cite{Dobre2015ACB}. 
    \item \textbf{BERTScore}: Leverages contextual embeddings from pre-trained transformer models to measure semantic similarity between generated and reference texts. Unlike n-gram-based metrics, BERTScore captures meaning and context, offering a robust evaluation for text generation tasks \cite{zhang2020bertscoreevaluatingtextgeneration}.
\end{itemize}

The reliability and faithfulness of generated text, particularly in applications requiring high accuracy, such as medical or financial domains is crucial. To identify inaccuracies, hallucination detection was conducted using Prometheus 2, a robust evaluation model designed for analyzing outputs of Large Language Models (LLMs) \cite{kim2024prometheus2opensource}. This framework helps evaluate three critical dimensions: 

\begin{itemize} 
    \item \textbf{Faithfulness}: Ensures that the generated content aligns with the source data and avoids unsupported claims \cite{madsen-etal-2022-evaluating, jacovi-goldberg-2020-towards}. 
    \item \textbf{Correctness}: Measures factual accuracy and checks for logical consistency in the output \cite{yao2023predictinggeneralizationperformancecorrectness, kim2024prometheus2opensource}. 
    \item \textbf{Fluency}: Evaluates the readability and linguistic quality of the text, ensuring it adheres to natural language norms \cite{suadaa-etal-2021-towards, Lee2023ASO}. 
\end{itemize}

\section{Models for Fine-tuning}
\label{sec:fine-tuning-models}

\begin{itemize}
    \item \textbf{LLaMA 2-Chat 7B} \cite{touvron2023llama}: LLaMA 2-Chat 7B is a fine-tuned variant of the LLaMA 2 series, optimized for dialogue applications. It employs an autoregressive transformer architecture and has been trained on a diverse dataset comprising 2 trillion tokens from publicly available sources. The fine-tuning process incorporates over one million human-annotated examples to enhance its conversational capabilities and alignment with human preferences for helpfulness and safety.

    \item \textbf{StructLM 7B} \cite{zhuang2024structlm}: StructLM 7B is a large language model fine-tuned specifically for structured knowledge grounding tasks. It utilizes the CodeLlama-Instruct model as its base and is trained on the SKGInstruct dataset, which encompasses a mixture of 19 structured knowledge grounding tasks. This specialized training enables StructLM to effectively process and generate text from structured data sources such as tables, databases, and knowledge graphs, making it robust in domain-specific text generation tasks.

    \item \textbf{Mistral 7B-Instruct} \cite{jiang2023mistral}: Mistral 7B-Instruct is an instruction fine-tuned version of the Mistral 7B model, designed to handle a wide array of tasks by following diverse instructions. It features a 32k context window and employs a Rope-theta of 1e6, without utilizing sliding-window attention. This configuration allows Mistral 7B-Instruct to perform effectively in multi-modal and domain-adapted text generation scenarios, achieving state-of-the-art performance in various benchmarks.
\end{itemize}

\section{Prometheus Evaluation} 
\label{appendix:Prometheus}

To evaluate model-based metrics, the Prometheus framework \cite{kim2024prometheus2opensource} was employed, utilizing structured prompts for three key evaluation criteria: fluency, correctness, and faithfulness. The primary framework leverages an Absolute System Prompt, which defines the role of the evaluator and ensures objective, consistent assessments based on established rubrics. This Absolute System Prompt, shown in Listing~\ref{code:ABS-System-Prompt}, forms the foundation for all evaluations across metrics.

\begin{lstlisting}[style=textstyle, frame = single, caption=Absolute System Prompt, label=code:ABS-System-Prompt]
You are a fair judge assistant tasked with providing clear, objective feedback based on specific criteria, ensuring each assessment reflects the absolute standards set for performance.
\end{lstlisting}
The task descriptions for evaluating fluency, correctness, and faithfulness share a similar structure, as shown in Listing~\ref{code:Task-description-Faithfulness},\ref{code:Task-description-fluency-correctness}. These instructions define the evaluation process, requiring detailed feedback and a score between 1 and 5, strictly adhering to a given rubric.


\begin{lstlisting}[style=textstyle, frame = single, caption=Task description used for evaluation of faithfulness, label=code:Task-description-Faithfulness]
###Task Description:
An instruction (might include an Input inside it), a response to evaluate, a reference answer that gets a score of 5, and a score rubric representing a evaluation criteria are given.
1. Write a detailed feedback that assess the quality of the response strictly based on the given score rubric, not evaluating in general.
2. After writing a feedback, write a score that is an integer between 1 and 5. You should refer to the score rubric.
3. The output format should look as follows: "Feedback: (write a feedback for criteria) [RESULT] (an integer number between 1 and 5)"
4. Please do not generate any other opening, closing, and explanations.
5. Only evaluate on common things between generated answer and reference answer. Don't evaluate on things which are present in reference answer but not in generated answer.
\end{lstlisting}

\subsection{Instructions for Evaluation}

Prometheus prompts are customized for each evaluation metric. Below are the specialized structures and rubrics for fluency, faithfulness, and correctness.

\paragraph{Faithfulness}
This metric ensures the generated response aligns with both the provided context and reference answers. The evaluation structure incorporates specific rubrics for relevance and information consistency.

\begin{lstlisting}[style=textstyle, frame = single, caption=Task description used for evaluation of fluency and correctness, label=code:Task-description-fluency-correctness]
###Task Description:
An instruction (might include an Input inside it), a response to evaluate, a reference answer that gets a score of 5, and a score rubric representing a evaluation criteria are given.
1. Write a detailed feedback that assess the quality of the response strictly based on the given score rubric, not evaluating in general.
2. After writing a feedback, write a score that is an integer between 1 and 5. You should refer to the score rubric.
3. The output format should look as follows: "Feedback: (write a feedback for criteria) [RESULT] (an integer number between 1 and 5)"
4. Please do not generate any other opening, closing, and explanations.
\end{lstlisting}

\begin{lstlisting}[style=textstyle, frame = single, caption=Prompt structured correctness, label=code:estructured-faithfulness]
###The instruction to evaluate:
Evaluate the fluency of the generated JSON answer.
###Context:
{Prompt}
###Existing answer (Score 5):
{reference_answer}
###Generate answer to evaluate:
{response}
###Score Rubrics:
"score1_description":"If the generated answer is not matching with any of the reference answers and also not having information from the context.",
"score2_description":"If the generated answer is having information from the context but not from existing answer and also have some irrelevant information.",
"score3_description":"If the generated answer is having relevant information from the context and some information from existing answer but have additional information that do not exist in context and also do not in existing answer.",
"score4_description":"If the generated answer is having relevant information from the context and some information from existing answer.",
"score5_description":"If the generated answer is matching with the existing answer and also having information from the context."}
###Feedback:
\end{lstlisting}

\paragraph{Fluency}
This metric evaluates the grammatical accuracy and readability of the generated response.

\begin{lstlisting}[style=textstyle, frame = single, caption=Prompt structured fluency, label=code:estructured-fluency]
###The instruction to evaluate: Evaluate 
the fluency of the generated JSON answer
###Response to evaluate: {response}
###Reference Answer (Score 5): 
{reference_answer}
###Score Rubrics:
"score1_description":"The generated JSON answer is not fluent and is difficult to understand.",
"score2_description":"The generated JSON answer has several grammatical errors and awkward phrasing.",
"score3_description":"The generated JSON answer is mostly fluent but contains some grammatical errors or awkward phrasing.",
"score4_description":"The generated JSON answer is fluent with minor grammatical errors or awkward phrasing.",
"score5_description":"The generated JSON answer is perfectly fluent with no grammatical errors or awkward phrase
###Feedback:
\end{lstlisting}

\paragraph{Correctness}
This metric assesses the logical accuracy and coherence of the generated response compared to the reference.

\begin{lstlisting}[style=textstyle, frame = single, caption=Prompt estructured correctness, label=code:estructured-correctness]
###The instruction to evaluate:
Your task is to evaluate the generated answer and reference answer for the query: {Prompt}
###Response to evaluate:
{response}
###Reference Answer (Score 5):
{reference_answer}
###Score Rubrics:
"criteria": "Is the model proficient in generate a coherence response",
"score1_description": "If the generated answer is not matching with any of the reference answers.",
"score2_description": "If the generated answer is according to reference answer but not relevant to user query.",
"score3_description": "If the generated answer is relevant to the user query and reference answer but contains mistakes.",
"score4_description": "If the generated answer is relevant to the user query and has the exact same metrics as the reference answer, but it is not as concise.",
"score5_description": "If the generated answer is relevant to the user query and fully correct according to the reference answer.

###Feedback: 
\end{lstlisting}

\section{Fine-tuning models}
\label{appendix:fine-tuning-models}

The prompts outlined below utilized for training eC-Tab2Text models (Listing \ref{code:Prompt Dataset 1}) and for the QTSumm dataset (Listing \ref{code:Prompt Dataset 2}).

\begin{lstlisting}[style=textstyle, frame = single, caption=Prompt structure for eC-Tab2Text, label=code:Prompt Dataset 1]
"Given following json that contains specifications of a product, generate a review of the key characteristics with json format. Follow the structure on Keys to write the Output: 

### Product: Product for JSON specifications

### Keys: Combination of the keys of the JSON reviews

### Output: reviews for JSON reviews accordingly to the keys"
\end{lstlisting}

\vspace{1cm}

\begin{lstlisting}[style=textstyle, frame = single, caption=Prompt structure for QTSumm, label=code:Prompt Dataset 2]
"Given following json that contains specifications of a product, generate a review of the key characteristics with json format. Follow the structure on Keys to write the Output: 
### Product: Column table of JSON specifications
### Keys: Column query of the dataset
### Output: Column summary of the dataset"
\end{lstlisting}

\section{eC-Tab2Text Data Formats} 
\label{Appendix: eC-Tab2Text Data Collection}

\begin{lstlisting}[style=jsonstyle, frame = single, caption=JSON Data Format Product specification, label=code:JSON-specs]
{
  "url": {
    "keys_specifications": [],
    "full_specifications": [
      "Launch Date": "Launch Date",
      "General": {
        "subcategories1": [
            "value1" ...
            ],
        "subcategories2": [
            "value1" ...
            ], ...
      },
      "Characteristic1": {
        "subcategories1": [
            "value1" ...
            ],
        "subcategories2": [
            "value1" ...
            ], ...
      },
      "Characteristic2": {
        "subcategories1": [
            "value1" ...
            ],
        "subcategories2": [
            "value1" ...
            ], ...
      }, ...
    ]
  },
}
\end{lstlisting}

\begin{lstlisting}[style=jsonstyle, frame = single, caption=JSON Data Format reviews, label=code:JSON-reviews]
{
  "url": {
    "text": {
      "Characteristic1": ["Description1"],
      "Characteristic2": ["Description2"], ...
    }
  }
}
\end{lstlisting}

\begin{figure*}[ht!]
    \centering
    \includegraphics[width=16cm]{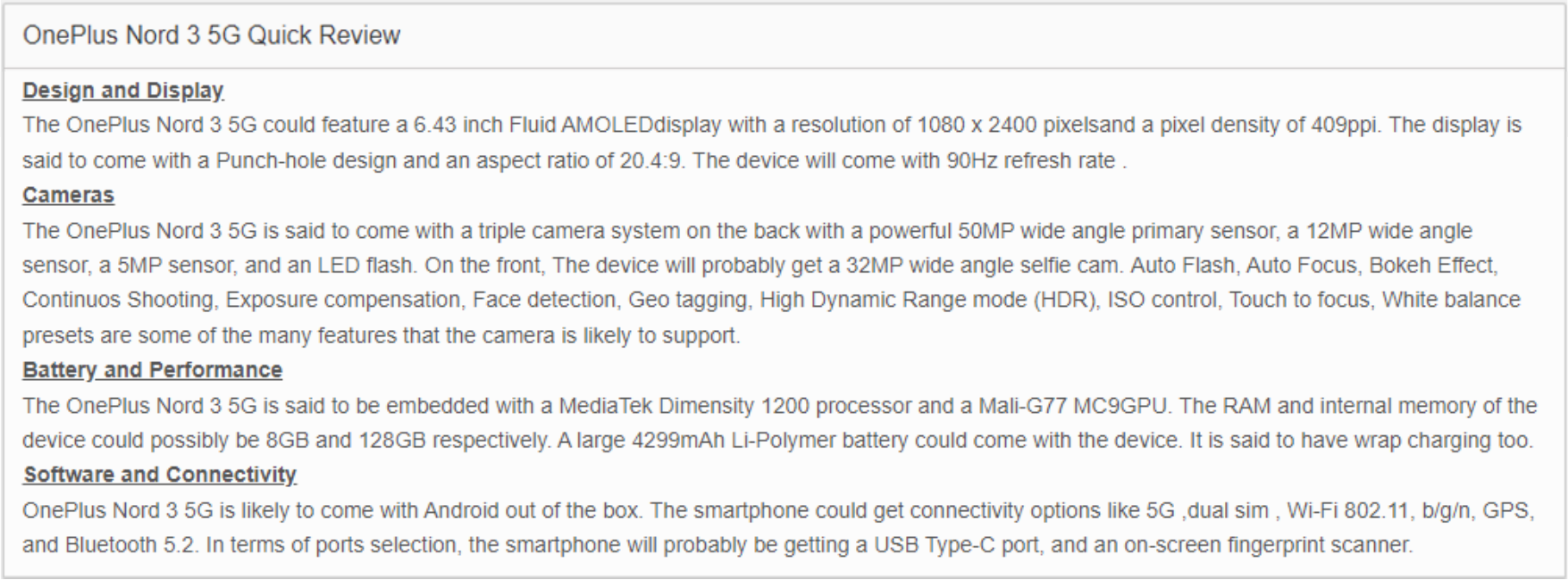}
    \caption{An illustration of sample output texts generated for user-specific queries based on structured input from product tables.}
    \label{fig:pricebaba-review-structure}
\end{figure*}

\begin{figure*}[ht!]
    \centering
    \includegraphics[width=16cm]{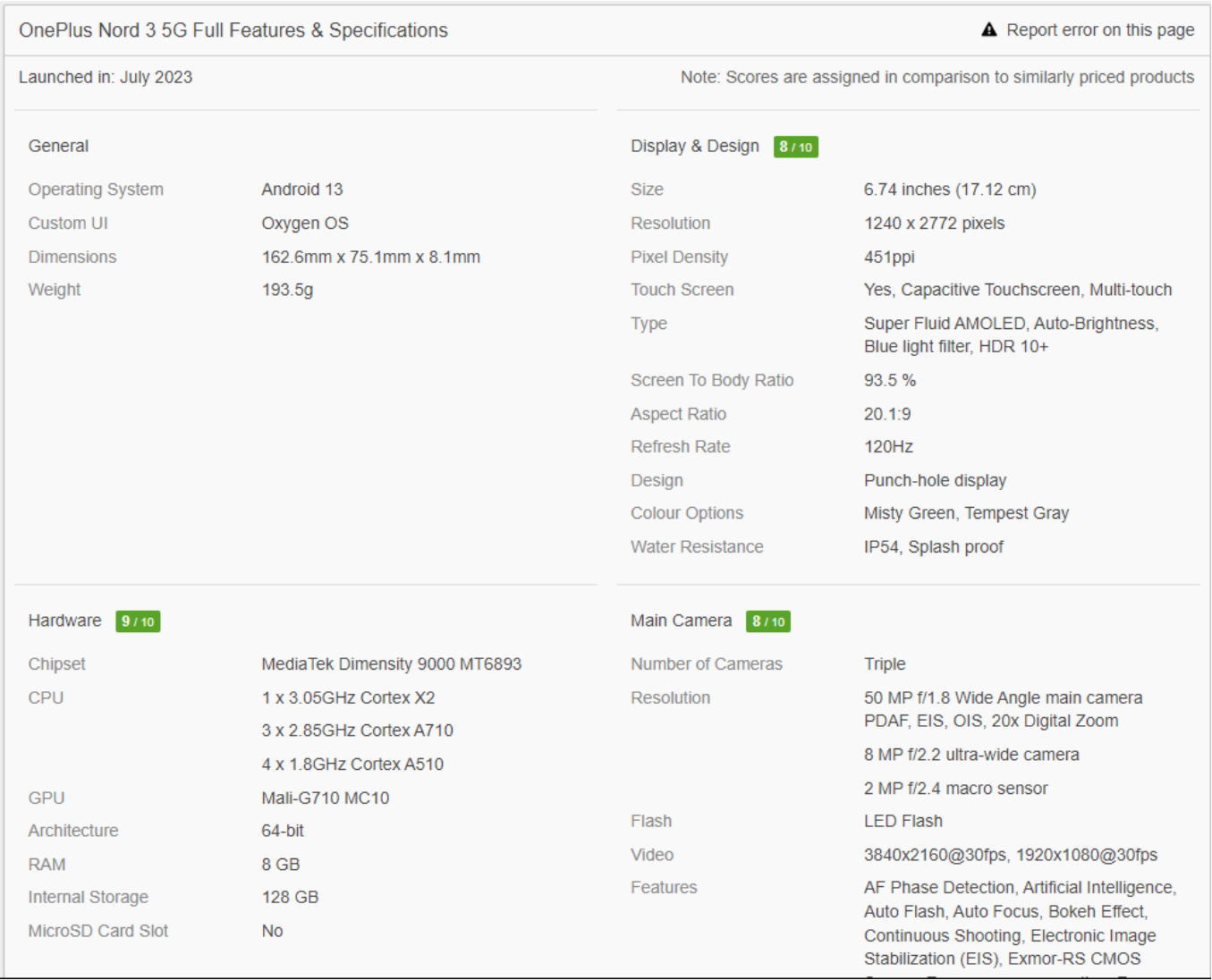}
    \caption{An example of a product specification table structure.}
    \label{fig:pricebaba-spec-structure}
\end{figure*}

\begin{figure*}
\begin{lstlisting}[style=reviews, frame = single, caption=Output generated with Mistral Instruct trained with eC-Tab2Text., label=code:JSON-Mistral-eC-Tab2Text]
{
    "Software and Connectivity": "Tecno Spark Go 2020 will run Android Go UI on top of Android 10 out of the box. For the connectivity, the smartphone will support features like Dual sim VoLTE (Micro-Micro), Wi-Fi 802.11, b/g/n, 3.5mm jack, Bluetooth 4.2, 10-pin charging, GPS, and USB 2.0.",
    "Design and Display": "The Tecno Spark Go 2020 will reportedly showcase a 6.52 inch TFTdisplay with a resolution of 720 x 1600 pixelsand a pixel density of 269ppi. The display will supposedly have a waterdrop notch with minimum bezels.",
    "Battery and Performance": "The Tecno Spark Go 2020 will be embedded with a MediaTek Helio A20 chipset and PowerVR GE8300 GPU. The device will come with 2GB RAM and 32GB of internal memory, which can be expanded up to 256GB. Fuelling the device will be a huge 5000mAh battery."
}
\end{lstlisting}
\end{figure*}
\begin{figure*}
\begin{lstlisting}[style=reviews, frame = single, caption=Output generated with Gemini1.5-flash., label=code:JSON-Gemini]
{
    "Cameras": {
        "Main Camera": {
            "Description": "Single 13 MP main camera with LED flash, Auto Flash, and Auto Focus. Offers digital zoom.",
            "Video": "Specifications not provided."
        },
        "Front Camera": {
            "Description": "Single 8 MP main camera.",
            "Video": "Specifications not provided."
        }
    },
    "Battery and Performance": {
        "Battery": {
            "Capacity": "4000 mAh Li-ion battery",
            "Charging": "No fast charging or wireless charging."
        },
        "Performance": {
            "Processor": "Quad-core 1.3GHz processor.  Specific chipset and GPU information not provided.",
            "RAM": "4GB RAM",
            "Storage": "64GB internal storage, expandable up to 64GB via microSD card."
        }
    },
    "Design and Display": {
        "Display": {
            "Size": "6.53 inches (16.59 cm) IPS LCD with 720 x 1560 pixels resolution (263 ppi).",
            "Design": "Waterdrop notch display."
        },
        "Other": {
            "Colors": "Available in Blue and Purple.",
            "Fingerprint Sensor": "No fingerprint sensor."
        }
    }
}
\end{lstlisting}
\end{figure*}
\begin{figure*}
\begin{lstlisting}[style=reviews, frame = single, caption=Output generated with GPT-4o-mini., label=code:JSON-GPT4]
{
    "Software and Connectivity": {
        "Operating System": "Android 10 (Q) with Android Go custom UI",
        "Network Compatibility": "Supports 2G, 3G, and 4G networks, making it suitable for use across various locations",
        "SIM Slots": "Dual SIM (Nano + Nano), allowing for flexible connectivity options",
        "Wi-Fi and Bluetooth": "Equipped with Wi-Fi 802.11 b/g/n and Bluetooth 4.2 for seamless wireless connections",
        "GPS": "Includes A-GPS and Glonass for accurate location tracking",
        "FM Radio and Charging Port": "Features FM Radio and uses a microUSB 2.0 charging port"
    },
    "Design and Display": {
        "Dimensions": "165.6mm x 76.3mm x 9.1mm, making it a reasonably sized smartphone",
        "Weight": "193g, providing a solid feel without being overly heavy",
        "Display Size": "6.52 inches with a TFT display that offers a resolution of 720 x 1600 pixels",
        "Pixel Density": "269ppi, delivering decent clarity for everyday usage",
        "Design": "Waterdrop notch design adds a modern touch; available in colour options of Aqua Blue and Ice Jadeite"
    },
    "Battery and Performance": {
        "Battery Capacity": "5000 mAh, ensuring long-lasting performance between charges",
        "Battery Type": "Li-ion, which is standard for smartphones in this range",
        "Charging Options": "No wireless charging and no fast charging feature, but provides adequate battery life throughout the day",
        "Hardware": {
            "Chipset": "MediaTek Helio A20",
            "RAM": "2 GB, sufficient for basic tasks and everyday applications",
            "Internal Storage": "32 GB, expandable up to 256 GB via microSD card, allowing for additional storage for apps and media"
        }
    }
}
\end{lstlisting}
\end{figure*}

\begin{figure*}
\begin{lstlisting}[style=reviews, frame = single, caption=Output generated with StructLM trained with eC-Tab2Text., label=code:JSON-StructLM]
{
    "Software and Connectivity": "Tecno Spark Go 2020 will run on Android Go UI on top of Android 10 out of the box. The smartphone will offer connectivity options like Dual SIM VoLTE, Wi-Fi 802.11, b/g/n, GPS, and Bluetooth 4.2. The device will also come with a rear-mounted fingerprint sensor and a 3.5mm headphone jack.",
    "Design and Display": "The Tecno Spark Go 2020 is expected to feature a 6.52 inch (16.56 cm) display with a screen resolution of 720 x 1600 pixels. The display will come with a waterdrop notch design and a screen-to-body ratio of 40.97 %. The device will come in Aqua Blue and Ice Jade colors.",
    "Battery and Performance": "The Tecno Spark Go 2020 will be powered by a MediaTek Helio A20 processor clocked at 2GHz and a PowerVR GE8300 GPU. The device will come armed with 2GB RAM and an internal storage capacity of 32GB, which can be expanded upto 256GB via a microSD card."
}
\end{lstlisting}
\end{figure*}

\begin{figure*}
\begin{lstlisting}[style=reviews, frame = single, caption=Output generated with Llama2 trained with eC-Tab2Text., label=code:JSON-Llama2]
{
    "Battery and Performance": "The Tecno Spark Go 2020 is anticipated to be powered by MediaTek\u2019s latest entry-level Helio A20quad-core chipset coupled with a PowerVR GE8300 GPU. That will again reportedly be paired with 2 GB RAM and an internal storage capacity of only 32 GB, which can further be expanded up to 256GB. Further, the device will reportedly pack in a 5000mAh Li-ion battery but won\u2019t support fast charging.",
    "Cameras": "The Tecno Spark Go 2020 is expected to house a dual-camera setup on the back with a 13MP (Digital Zoom) camera as the primary sensor. Besides, there will also be a 2MP depth sensor onboard. On the front, the smartphone will supposedly feature an 8MP selfie shooter. There will also likely be a bunch of camera features such as Artificial Intelligence,Auto Flash,Auto Focus,Bokeh Effect,Continuos Shooting,Exposure compensation,Face detection,Geo tagging,High Dynamic Range mode (HDR),ISO control,Touch to focus,White balance presets.",
    "Design and Display": "The Tecno Spark Go 2020 will reportedly feature a 6.52 inch TFT panel tipped with a resolution of 720 x 1600 pixels. The pixel density will supposedly max out at 269ppi. The bezel-less display is further anticipated to boast a waterdrop notch design to furnish an immersive viewing experience."
}
\end{lstlisting}
\end{figure*}

\end{document}